\documentclass[runningheads]{llncs}

\usepackage{hyperref}
\usepackage{marvosym}

\usepackage{graphicx}
\usepackage{amsmath}
\usepackage{amssymb}
\usepackage{booktabs}
\usepackage{hhline}
\usepackage{multirow}
\usepackage{multicol}
\usepackage{colortbl}
\usepackage[table,xcdraw]{xcolor}
\usepackage[a4paper, top=5.2cm, bottom=5.2cm, left=4.4cm, right=4.4cm]{geometry}

\hypersetup{
hidelinks,
colorlinks=true,
linkcolor=red,
citecolor=green
}
\usepackage[capitalize]{cleveref}
\crefname{section}{Sec.}{Secs.}
\Crefname{section}{Section}{Sections}
\Crefname{table}{Table}{Tables}
\crefname{table}{Tab.}{Tabs.}

\usepackage{graphicx}
\begin{document}
\title{PlantDet: A benchmark for Plant Detection
in the Three-Rivers-Source Region}
\author{Huanhuan Li\inst{1}\orcidID{0009-0004-1825-5025} \and
Yu-an Zhang\inst{1}\thanks{Corresponding author.}\orcidID{0000-0002-7450-9876} \and
Xuechao Zou\inst{1}\thanks{Huanhuan Li and Xuechao Zou have contributed equally to this work.}
\and
Zhiyong Li\inst{1}
\and
Jiangcai Zhaba\inst{2}
\and
Guomei Li\inst{2}\and
Lamao Yongga\inst{2}
}
\authorrunning{F. Author et al.}

\institute{Department of Computer Technology and Applications, Qinghai University, Xining, China \url{https://www.qhu.edu.cn/}\and
Forestry and Grassland Comprehensive Service Center of Yushu Prefecture, Yushu, China\\
}
\maketitle              
\begin{abstract}
The Three-River-Source region is a highly significant natural reserve in China that harbors a plethora of botanical resources. To meet the practical requirements of botanical research and intelligent plant management, we construct a dataset for \textbf{P}lant detection in the \textbf{T}hree-\textbf{R}iver-\textbf{S}ource region (PTRS). It comprises 21 types, 6965 high-resolution images of 2160$\times$3840 pixels, captured by diverse sensors and platforms, and featuring objects of varying shapes and sizes. The PTRS presents us with challenges such as dense occlusion, varying leaf resolutions, and high feature similarity among plants, prompting us to develop a novel object detection network named PlantDet. This network employs a window-based efficient self-attention module (ST block) to generate robust feature representation at multiple scales, improving the detection efficiency for small and densely-occluded objects. Our experimental results validate the efficacy of our proposed plant detection benchmark, with a precision of 88.1\%, a mean average precision (mAP) of 77.6\%, and a higher recall compared to the baseline. Additionally, our method effectively overcomes the issue of missing small objects. 
 
\end{abstract}
\section{Introduction}
The Three-Rivers-Source region is located in the hinterland of the Qinghai-Tibet Plateau, in the southern part of Qinghai Province. It is the largest nature reserve in China, containing extremely rich wild plant resources. 
In recent years, the conservation of flora and fauna in the Three-Rivers-Source region has become a focus of attention. However, due to its remote geographical location, underdeveloped information technology, people's awareness of vegetation protection in the Three-Rivers-Source region is relatively low. Therefore, conducting a survey of plant resources in the Three-Rivers-Source region, especially in plant detection, is of great significance for achieving intelligent plant management and protection.

In recent years, with the rapid development of artificial intelligence and computer vision, many convolutional neural network models based on deep learning,such as AlexNet\cite{alexnet}, ResNet\cite{resnet}, and VGGNet\cite{vggnet}, have emerged. They have propelled the development of object detection algorithms. The introduction of algorithms such as SSD\cite{ssd}, YOLO series\cite{yolov1,yolov2,yolov3,yolov4,yolov6,yolov7}, 
and the algorithms based on the R-CNN\cite{rcnn}, has expanded the promotion and application of object detection in the agricultural field. Numerous experimental results have shown that algorithmic models based on convolutional neural networks perform well in plant recognition research. Therefore, utilizing artificial intelligence and deep learning technology to detect plants in the Three-Rivers-Source region is feasible.

In essence, we have made the following contributions:
\begin{itemize}
    \item We collected 6965 plant images of 21 categories from the Three-River-Source region, and manually annotated them to establish a large-scale dataset called PTRS for plant detection. This dataset lays the foundation for precise and modern plant detection in the Three-River-Source region.
    \item We proposed a novel object detection benchmark called PlantDet on PTRS to tackle the challenges of uneven leaf sizes and high feature similarity of diverse plant species. 
    This method consists of three parts: Backbone, Neck, and Head. We introduced an efficient self-attention module based on sliding windows to enhance the feature extraction ability of the backbone and obtained robust feature representation of different scales through efficient feature fusion strategies.
    \item Experimental results on PTRS demonstrated that our benchmark (PlantDet) surpasses the baseline (YOLOv5), achieves a precision of 88.1\% and mAP of 77.6\%, and mitigates the problem of missed detection and false positives for small objects.
\end{itemize}

\section{Related work}
\subsection{Object Detection}
Compared to image classification, object detection not only identifies the category of various objects in the image but also determines their location. Object detection can be divided into two types : one is the two-stage algorithm represented by R-CNN\cite{rcnn}. The principle of such methods is to generate candidate boxes, search for prospects, and adjust bounding boxes through specialized modules.Although this candidate region-based detection method has relatively high accuracy, it runs slowly and does not meet the demand for real-time detection.

To tackle the crucial issue of slow detection speed, one-stage object detection algorithms such as SSD and YOLO series algorithms have emerged. They consider the detection task as a regression problem and directly classify and locate objects in the image through a single neural network.
Due to the usage of a single network, they are relatively faster and can meet the real-time detection requirements in the industry. The YOLO series of algorithms have been widely applied in the agricultural field such as detecting diseases and pests\cite{diseases}, maturity\cite{maturity}, and growth stages\cite{growth}, among others. 

\subsection{Visual Transformer}

In 2017, the Google research team proposed the transformer architecture based on the self-attention mechanism, which achieved tremendous success in the field of natural language processing. The rapid development of the transformer in natural language processing has attracted widespread attention in the field of computer vision. The advantage of a transformer lies in its explicit modeling of long-range dependencies between contextual information, 
so many researchers have attempted to apply the transformer to computer vision in order to enhance the overall perceptual ability of images. In 2020 Carion et al.\cite{detr} proposed the first end-to-end transformer-based object detection model. That same year, the proposal of the image classification model ViT\cite{vit} led to the rapid development of visual transformers.

Today, the visual transformer is widely used in various computer vision fields, such as image classification, object detection, image segmentation, and object tracking. So far, many algorithm models based on the visual transformer have emerged: 1) Transformer-based object detection and segmentation models, such as Swin Transformer\cite{swin} and Focal Transformer\cite{focal-transformer} which replace CNN-based backbone networks for feature extraction and combine classic object detection and segmentation networks to complete detection and segmentation tasks; 2) Transformer-based object tracking tasks, such as TrSiam\cite{trsiam} for single-object tracking tasks, TransTrack\cite{transtrack} for multi-object tracking. The rapid development of the transformer in computer vision is mainly due to its ability to extract the relevance of contextual information to obtain global receptive fields, which improves the performance of the model compared to CNN-based models.

\section{Method}

\subsection{Overall Pipeline}
Plant detection is an application of object detection technology in botany. A deep learning-based task takes an image with plants as input and outputs the plant's category and bounding box location of its leaves. The Three-River-Source region has diverse flora, to achieve real-time detection, we use YOLOv5 as the baseline for the plant detection pipeline.

\begin{figure}[ht]
\centering
\includegraphics[width=\linewidth]{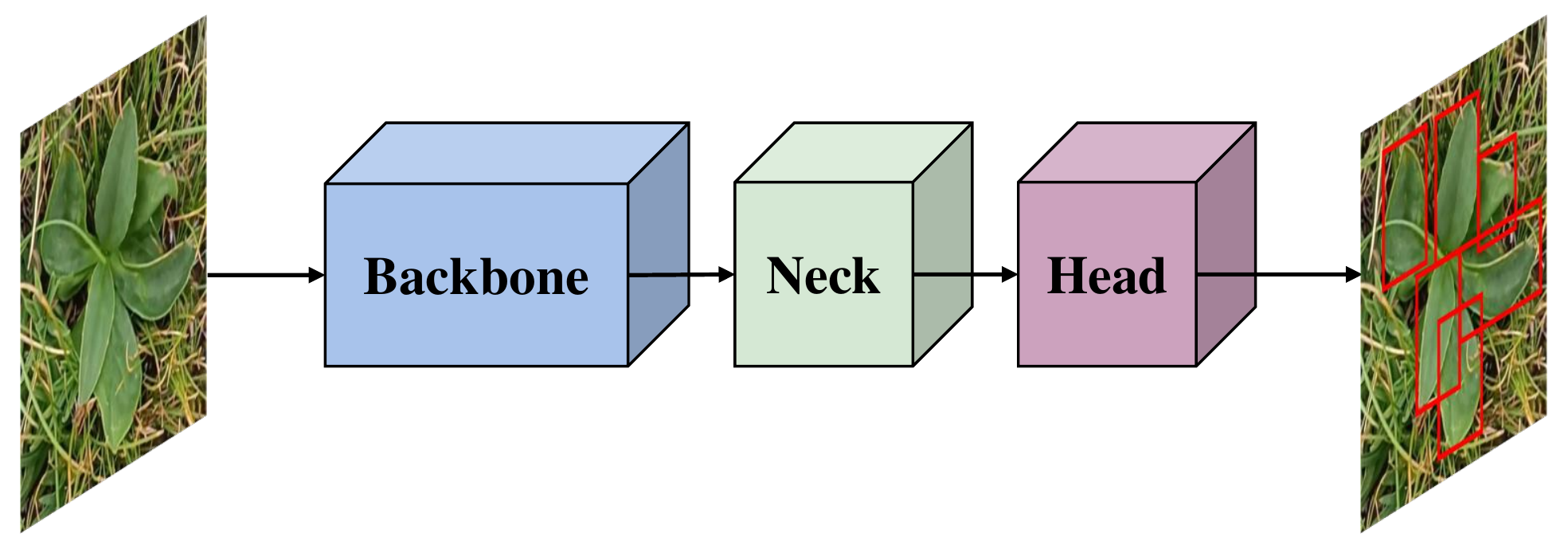}
\caption{A pipeline  of our one-stage plant detection methods (like the Y0LO series). 
}
\label{fig:pipeline}
\end{figure}

Having an efficient model structure is one of the most critical issues in designing a real-time object detector. Our proposed method, PlantDet, uses CSPDarkNet and CSPPAFPN composed of the same building units for multi-scale feature fusion, and finally inputs the features into different detection heads. The overall model structure of PlantDet is shown in Fig.~\ref{fig:pipeline}. PlantDet consists of three parts: 1) Backbone: it mainly performs feature extraction in the main part and effectively extracts crucial feature information of the feature map through downsampling; 2) Neck: this part consists of FPN\cite{fpn} and PAN\cite{pan}, respectively performing upsampling and downsampling to achieve the transmission of object feature vectors of different scales and fusion of multiple feature layers; 3) Head: which is made up of three multi-scale detectors and performs object detection on feature maps of different scales using grid-based anchors.

\subsection{Detection Backbone}
\begin{figure*}[ht]
\centering
\includegraphics[width=\linewidth]{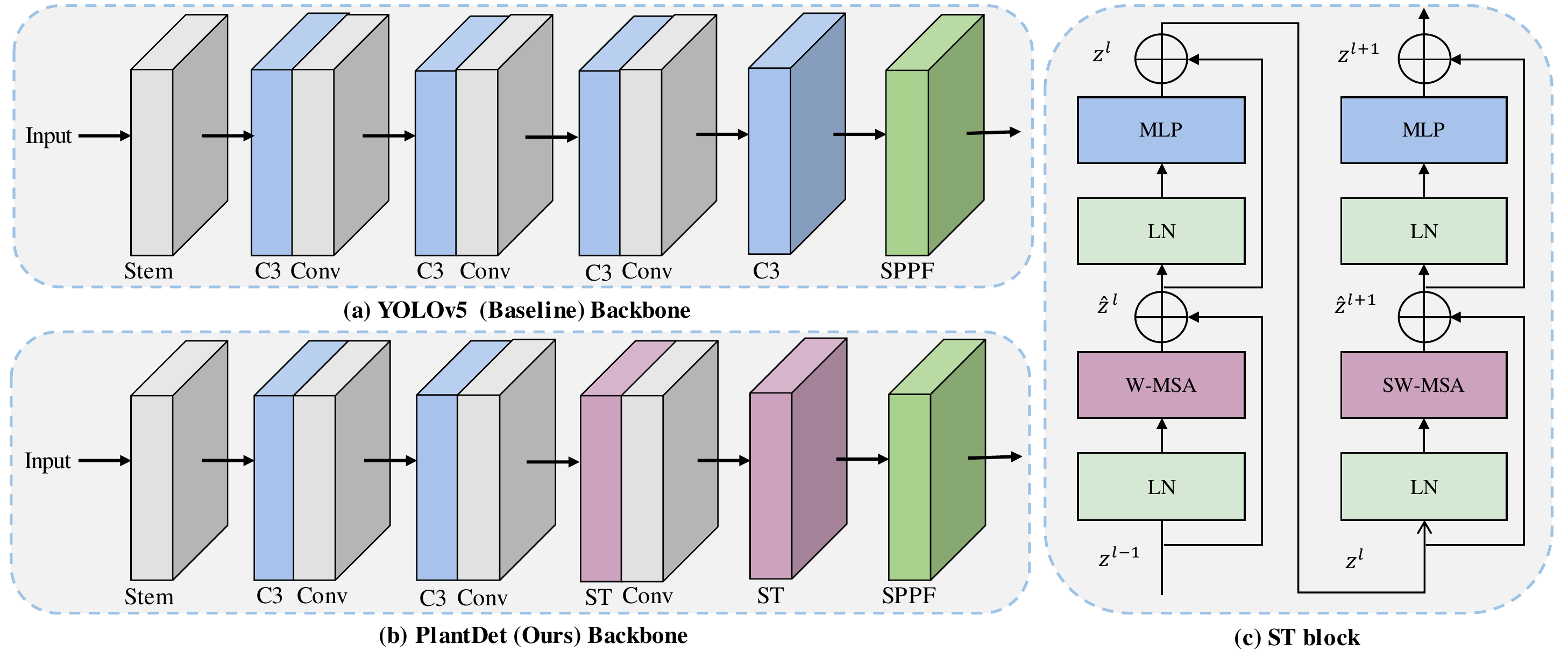}
\caption{Backbone of our proposed PlantDet and details of the ST block. (a) Structure of the original YOLOv5's backbone. (b) Structure of our proposed PlantDet's backbone. (c) Details of the ST block in PlantDet. 
The specially designed ST block is used for extracting global contextual information, mainly composed of W-MSA and SW-MSA, for information exchange within and between windows, respectively.
}
\label{fig:backbone}
\end{figure*}

We use YOLOv5 as the baseline for the plant detection pipeline. YOLOv5 consists of Input, Backbone, Neck and Head. The backbone (refer to Fig.~\ref{fig:backbone}(a)) mainly includes C3 and SPPF , where C3 consists of a CBS layer with x residual connections for Concat operation, which improves the feature extraction ability and retains more feature information. SPPF first performs the extracted feature map for multiple maximum pooling operations, and then the results after each maximum pooling are summed for Contact operation, i.e., feature fusion.

In response to the issue of difficult detection caused by varying distributions of leaf sizes in different plants, severe occlusion, and high feature overlap,we have introduced a sliding window module based on self-attention and embedded it into the backbone (refer to Fig.~\ref{fig:backbone}) to obtain a robust feature representation with multi-scale resolution. 

Specifically, we have re-designed the C3 module in the YOLOv5 backbone, which has the most significant impact on feature extraction. 
The C3 module primarily acquires feature representation through two parallel convolution branches and introduces residual connections, but does not consider modeling global contextual information. Therefore, we have introduced a self-attention module named "ST block" (refer to Fig.~\ref{fig:backbone}(c)) to obtain a global receptive field and more robust representation. 

The ST block includes sliding window operation with hierarchical design. It consists of LayerNorm and a shifted window-based MSA with two layers of MLP. Firstly, input features are normalized using Layer Normalization (LN) to expedite model convergence. Subsequently, global feature representation is obtained through the multi-head self-attention mechanism. Furthermore, the features are further enhanced and their expression ability is strengthened through the use of MLP. Finally, residual connection is employed for feature reuse. In addition, a window mechanism is utilized to reduce the additional overhead resulting from the calculation of self-attention matrices.

As is well known, Convolutional Neural Networks (CNNs) perform exceptionally well in local feature extraction due to their inductive bias, while transformer networks based on self-attention mechanisms are effective in modeling long-range global contextual information. Taking into account the superiority of both convolutional and self-attention mechanisms, we have designed a robust backbone feature extractor for plant detection, as shown in Fig.~\ref{fig:backbone}(b). It consists of two C3 modules for local feature extraction and two ST blocks for global feature extraction. Finally, the SPPF module fuses the features extracted by both modules to obtain a robust feature representation.

\subsection{Loss Function}

The task of object detection involves the regression of bounding boxes in addition to classification. Consequently, the training loss function comprises three parts: 1) bounding box regression loss; 2) confidence prediction loss; 3) category prediction loss.
These three loss functions are jointly optimized to achieve the goal of object detection
\begin{equation}
\mathcal{L}=\lambda_{1}\mathcal{L}_{reg}+\lambda_{2}\mathcal{L}_{obj}+\lambda_{3}\mathcal{L}_{cls},
\end{equation}
where $\lambda_{1},\lambda_{2},\lambda_{3}$  represent the weights of the three loss functions, respectively.

\paragraph{\textbf{Bounding Box Regression Loss.}} To account for the large variation in scale among different plant leaves and to balance
the impact of objects of different sizes on detection performance, we use the Complete-IoU(CIoU)\cite{ciou} to calculate the bounding box regression loss
\begin{equation}
    \mathcal{L}_{reg}=CIoU=IoU-\frac{\rho^{2}}{c^{2}}-\alpha v ,
\end{equation}
%
Where $\rho$, $c$, and $v$ respectively represent the distance,the diagonal length and the similarity of aspect ratio between the centers of the predicted and ground-truth bounding boxes, and $\alpha$ represents the impact factor of $v$.

\paragraph{\textbf{Confidence Loss.}}The loss function for confidence prediction is computed by matching positive and negative samples. Firstly, it involves the predicted confidence within the bounding box. Secondly, it uses the Intersection over Union (IoU) value between the predicted bounding box and its corresponding ground-truth bounding box as the ground-truth value. These two values are then used to calculate the final loss for the confidence prediction, which is obtained through binary cross-entropy
\begin{equation}
    \mathcal{L}_{obj}(p_{o}, p_{iou}) = BCE_{obj}^{sig} (p_{o}, p_{iou};w_{obj}),
\end{equation}
where $p_{o}$ and $p_{iou}$ represent the predicted confidence and ground truth confidence, respectively, $w_{obj}$ demonstrates the weight of positive samples.

\paragraph{\textbf{Classification Loss.}} The category prediction loss is similar to the confidence loss. It involves predicting the category score within the bounding box and using the ground-truth one-hot encoding of the category for the corresponding ground-truth bounding box. The category prediction loss is computed using the following formula 
\begin{equation}
   \mathcal{L}_{cls}(c_{p}, c_{gt}) = BCE_{cls}^{sig} (c_{p}, c_{gt};w_{cls}),
\end{equation}
where $c_{p}$ and $c_{gt}$ represent the predicted values for the corresponding categories.

\section{Experiments}

\subsection{Dataset}

\begin{figure*}[ht]
\centering
\includegraphics[width=\linewidth]{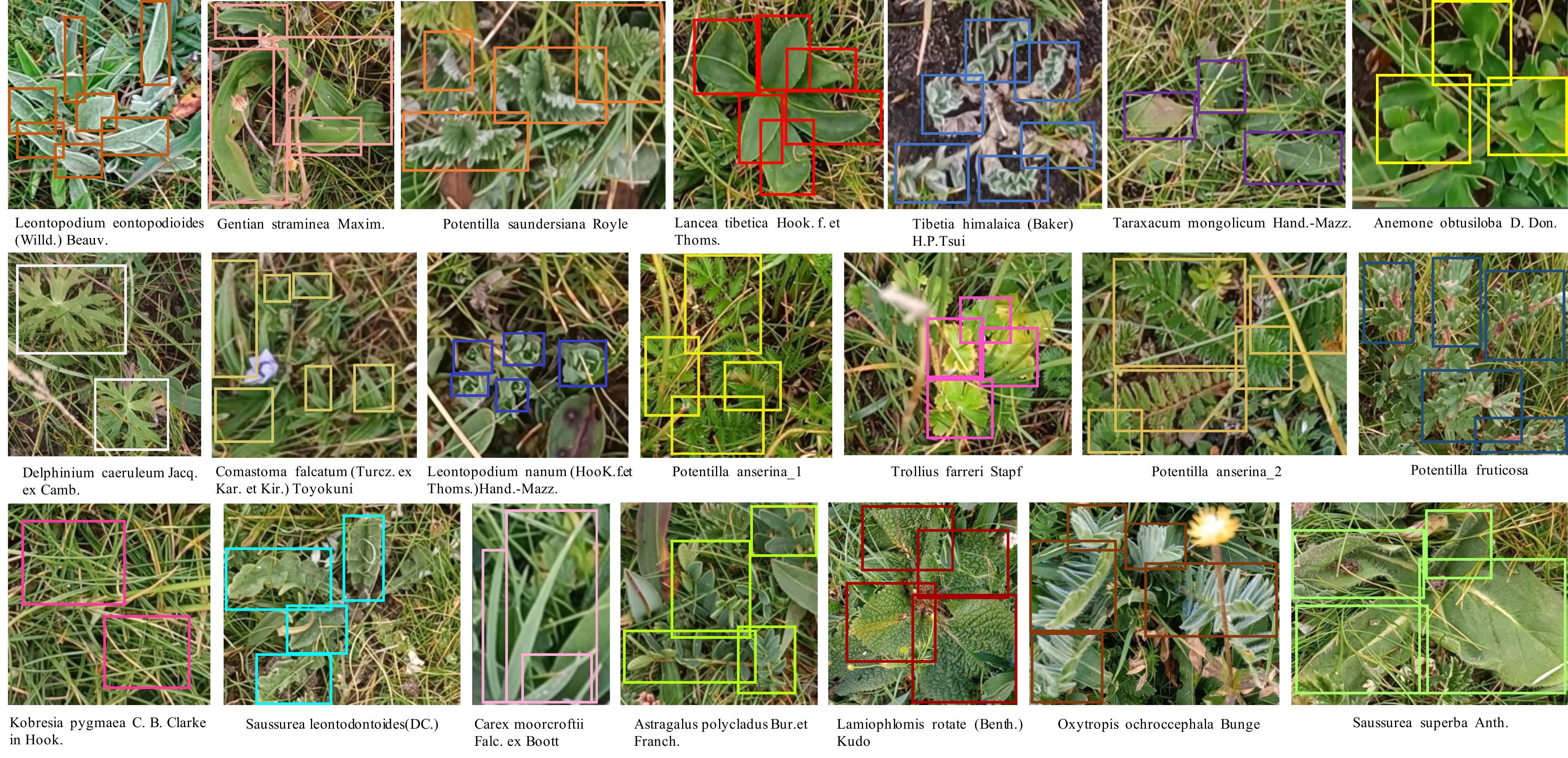}
\caption{Samples and corresponding Latin names of our dataset for \textbf{P}lant detection in the \textbf{T}hree-\textbf{R}iver-\textbf{S}ource region (PTRS). 
}
\label{fig:samples}
\end{figure*}

\paragraph{\textbf{Data Collection.}} The research object of this experiment is the vegetation in the grassland plots distributed in the Three-River-Source region. The plant species image data were taken between July and August 2022 using a handheld camera, approximately 20 centimeters away from the plot, and recorded at a certain speed. After processing, 6965 grassland images were obtained, involving 21 plant species. The plant images involved in the experiment and their corresponding Latin names are shown in Fig.~\ref{fig:samples}. These plants were all identified by experienced experts in the field.

\paragraph{\textbf{Data Annotations.}} 6965 images of 21 plant species involved in this experiment were annotated by experienced experts in the field. Initially, the Make Sense({\url{https://www.makesense.ai/}}) labeling tool provided by YOLO was utilized to generate label files containing information about plant categories and target plant coordinates. Subsequently, the above data was organized into VOC format datasets for \textbf{P}lant detection in the \textbf{T}hree-\textbf{R}iver-\textbf{S}ource region (PTRS). Finally, PTRS was divided into training, validation, and testing sets in an 8:1:1 ratio. In addition, comparing our dataset PTRS with other plant detection datasets, the detailed comparison results of these existing datasets are shown in Table \ref{tab:0}.
\begin{table}[ht]
\centering
\caption{Comparison among PTRS (Ours) and other plant detection datasets in agriculture. ``-" indicates that this metric is not revealed in the original paper.
}
\label{tab:0}
\begin{tabular}{cccccc}
\hline
Dataset                            & Annotation way & Classes & Instances                           & Images                     & Image Size                     \\ \hline
AT\cite{at} & Oriented Bounding Box    & 1               & $~$1000                               & 1000                       & 410*410 \\
GHLD\cite{ghld}         & Horizontal Bounding Box  & 1               & \_                                  & 300                        & 416*416 \\
TDAP\cite{tdap}                      & Horizontal Bounding Box  & 1               & \_                                  & 5000                       & \_                             \\
TFP\cite{tfp}                       & Oriented Bounding Box    & 1               & \_                                  & 814 & \_                             \\
GPSD\cite{gpsd}                               & Horizontal Bounding Box  & 4               & \_                                  & 1200                       & \_                             \\ \hline
\textbf{PTRS (Ours)}                      & Horizontal Bounding Box  & \textbf{21}     & \multicolumn{1}{l}{\textbf{122300}} & \textbf{6965}              & \textbf{2160*3840}             \\ \hline
\end{tabular}
\end{table}

\subsection{Implementation Details}

\paragraph{\textbf{Training Settings.}} The important training parameters for the model in this experiment were set as follows: training epoch of 300, uniform resizing of input images to 640$\times$640 resolution, 
training batch-size of 32, an initial learning rate of 0.01 with Stochastic Gradient Descent (SGD) optimizer. The model was trained on a device with a GPU of 1xNVIDIA A100 and 80GB memory, and the deep learning framework PyTorch was used for implementation.

\paragraph{\textbf{Evaluation Metrics.}} In these experiments, Precision (P), Recall (R), and mean of Average Precision (mAP@0.5) are used as evaluation metrics. 

\subsection{Ablation Studies}

\paragraph{\textbf{Transformer Backbone.}} To investigate the efficacy of self-attention mechanisms and determine the optimal mechanism applicable to plant detection, we conducted experiments using YOLOv5 as the baseline as shown in Table~\ref{tab:1}.


\begin{table}[ht]
\centering
\caption{Ablation experiments of self-attention mechanisms.}
\label{tab:1}
\begin{tabular}{cccc}
\hline
Self-Attention      & Precision   & Recall & mAP@0.5  \\ \hline
Baseline       & 88.0 & 71.9   & 76.6 \\
Baseline+MSA   & 86.1 & 66.2   & 72.1 \\
\rowcolor[RGB]{176,224,230} Baseline+W-MSA & \textbf{88.1} & \textbf{72.9}   & \textbf{77.6} \\ \hline
\end{tabular}

\end{table}

The MSA represents the original implementation of self-attention, whereas the W-MSA is a window-based self-attention mechanism. The experimental results demonstrate that compared to the model combined with MSA, the combination of W-MSA module yields better results on the PTRS dataset. Specifically, the Precision, Recall, and mAP were improved by 2.0\%, 6.7\%, and 5.5\%, respectively. This improvement is attributed to the fact that the W-MSA 
is constructed based on the image resolution hierarchy, which 
not only achieves feature connections across different windows but also enhances information exchange among different windows, allowing for the extraction of more effective multi-scale feature information to exhibit superior detection performance. 

\paragraph{\textbf{Strategy for Combining Global and Local Modules.}} In the original YOLOv5, the feature extraction network of the backbone consists of four C3 modules. We conducted ablation experiments to explore the impact of different module combination strategies (C3 and ST block) on the detection results, and the results are shown in Table~\ref{tab:2}.

\begin{table}[ht]
\centering
\caption{Ablation experiments of module combination strategies.}
\label{tab:2}
\begin{tabular}{ccccc}
\hline
\multicolumn{2}{c}{Number of Module} & \multirow{2}{*}{Precision} & \multirow{2}{*}{Recall} & \multirow{2}{*}{mAP@0.5} \\ \hhline{--}
C3                & ST block          &                     &                         &                      \\ \hline  
0                 & 4                 & 87.3                & 70.8                    & 75.9                 \\
1                 & 3                 & 84.3                & 72.5                    & 75.8                 \\
\rowcolor[RGB]{176,224,230} \textbf{2}        & \textbf{2}        & \textbf{88.1}       & \textbf{72.9}           & \textbf{77.6}        \\
3                 & 1                 & 85.7                & 72.3                    & 76.0                 \\
4                 & 0                 & 88.0                & 71.9                    & 76.6                 \\ \hline
\end{tabular}
\end{table}

The results indicate that the best performance in feature extraction is achieved by using two C3 modules and two ST blocks in the backbone. This is because the C3 module based on the convolutional network can extract local features, while the ST block based on self-attention can extract global features, and the fusion of the two types of features can obtain a more robust feature representation. Therefore, we use two C3 modules and two ST blocks for feature extraction, aiming to improve model performance while minimizing model parameters and computation time complexity.

\section{Comparison with the State-of-the-Arts}

\paragraph{\textbf{Quantitative Comparison.}} We conducted experiments to quantitatively compare PlantDet with currently popular object detection algorithms on our self-made PTRS dataset. The results are shown in Table~\ref{tab:4}. The results indicate that comparing to the baseline YOLOv5, the Recall and mAP of PlantDet increased by 1\%, and achieves SOTA results. The outstanding performance of PlantDet is due to the robust detection backbone we have proposed, which integrates global and local information to obtain a more robust multi-scale feature representation. In addition, the numerical evaluation results of Precision, Recall and mAP of baseline (YOLOv5) and PlantDet on the PTRS dataset are shown in 
Table ~\ref{tab:6}. 

\begin{table}[ht]
\centering
\caption{Quantitative comparison between our and existing models on the dat-\\aset.}
\label{tab:4}
\begin{tabular}{cccc}
\hline
Mothods             & Precision            & Recall        & mAP@0.5           \\ \hline
SSD~\cite{ssd}                    &46.6           & 18.6             & 48.9            \\
FCOS~\cite{fcos}                 & -          & 71.8          & 57.4          \\
CornerNet\cite{cornernet}            & 11.0       & 51.9          & 38.1          \\
Fast R-CNN~\cite{fast-r-cnn}               & -           & 40.0           & 56.3          \\
YOLOF~\cite{yolof}                 & -          & 69.7          & 54.6          \\
YOLOv7~\cite{yolov7}                 & 84.9          & 72.7          & 76.0          \\
YOLOv5                 & 88.0          & 71.9          & 76.6          \\
\rowcolor[RGB]{176,224,230}\textbf{PlantDet (Ours)} & \textbf{88.1} & \textbf{72.9} & \textbf{77.6} \\ \hline
\end{tabular}
\end{table}

\begin{table}[ht]
\centering
\caption{Numerical results of YOLOv5 and our PlantDet in 21 categories of our PTRS dataset. Size represents the size of plant leaves, obtained by calculating the bounding box size of all class instances. It can be easily observed that PlantDet enhances the detection performance of small and medium-sized targets, and has a superior effect.
}
\begin{tabular}{cccccccc}
\hline
\multirow{2}{*}{Plant} & \multirow{2}{*}{Size} & \multicolumn{3}{c}{YOLOv5 (Baseline)} & \multicolumn{3}{c}{PlantDet (Ours)}            \\ \cmidrule(r){3-5} \cmidrule(r){6-8}
                       &                       & Precision         & Recall       & mAP@0.5       & Precision            & Recall        & mAP@0.5           \\ \hline
01                     & Small                     & 86.9       & 66.6         & 73.6      & 85.4          & \textbf{65.4} & 72.8          \\
02                     & Medium                     & 88.9       & 77.1         & 82.3      & 88.9          & 76.7          & 82.1          \\
03                     & Medium                     & 89.2       & 74.9         & 80.8      & \textbf{89.5} & \textbf{74.9} & \textbf{80.8} \\
04                     & Large                     & 90.7       & 75.7         & 81.5      & 88.6          & \textbf{76.3} & \textbf{81.8} \\
05                     & Medium                     & 85.5       & 66.4         & 70.3      & 82.4          & 63.8          & 70.2          \\
06                     & Medium                     & 87.7       & 73.6         & 78.0      & 82.5          & \textbf{74.3} & 77.2          \\
07                     & Small                     & 90.0       & 73.3         & 77.8      & \textbf{92.9} & 68.0          & 75.0          \\
08                     & Large                     & 95.5       & 77.9         & 81.1      & \textbf{96.2} & 77.0          & 79.1          \\
\rowcolor[RGB]{176,224,230}09                     & Small                     & 82.8       & 67.2         & 69.8      & \textbf{84.2} & \textbf{67.5} & \textbf{71.5} \\
\rowcolor[RGB]{176,224,230}10                     & Small                     & 87.7       & 72.7         & 81.3      & \textbf{98.8} & \textbf{90.9} & \textbf{90.6} \\
\rowcolor[RGB]{176,224,230}11                     & Medium                     & 93.5       & 76.3         & 81.1      & 92.3          & 75.2          & \textbf{81.9} \\
\rowcolor[RGB]{176,224,230}12                     & Small                     & 89.7       & 76.0         & 84.5      & 89.5          & \textbf{84.0} & \textbf{88.2} \\
\rowcolor[RGB]{176,224,230}13                     & Medium                     & 86.8       & 69.7         & 74.6      & 85.0          & \textbf{69.7} & 73.8          \\
\rowcolor[RGB]{176,224,230}14                     & Medium                     & 91.1       & 77.1         & 81.0      & 90.0          & \textbf{77.9} & \textbf{81.8} \\
\rowcolor[RGB]{176,224,230}15                     & Small                     & 77.2       & 53.4         & 57.6      & 73.2          & \textbf{55.1} & \textbf{59.6} \\
\rowcolor[RGB]{176,224,230}16                     & Small                     & 93.7       & 77.4         & 81.4      & 93.6          & \textbf{80.2} & \textbf{85.4} \\
\rowcolor[RGB]{176,224,230}17                     & Small                     & 82.6       & 68.2         & 69.7      & 81.7          & 68.1          & \textbf{70.7} \\
\rowcolor[RGB]{176,224,230}18                     & Small                     & 85.7       & 69.7         & 73.7      & \textbf{86.1} & \textbf{69.7} & \textbf{75.0} \\
19                     & Large                     & 79.8       & 66.7         & 66.0      & 80.4          & 61.9          & 63.7          \\
20                     & Medium                     & 93.3       & 86.0         & 92.7      & \textbf{97.3} & \textbf{86.0} & 89.7          \\
21                     & Large                     & 88.9       & 63.0         & 70.3      & \textbf{91.8} & \textbf{68.5} & \textbf{79.6} \\ \hline
Avg.                   & -                     & 88.0       & 71.9         & 76.6      & \textbf{88.1} & \textbf{72.9} & \textbf{77.6} \\ \hline
\end{tabular}

\label{tab:6}
\end{table}

\begin{figure*}[ht]
\centering
\includegraphics[width=\linewidth]{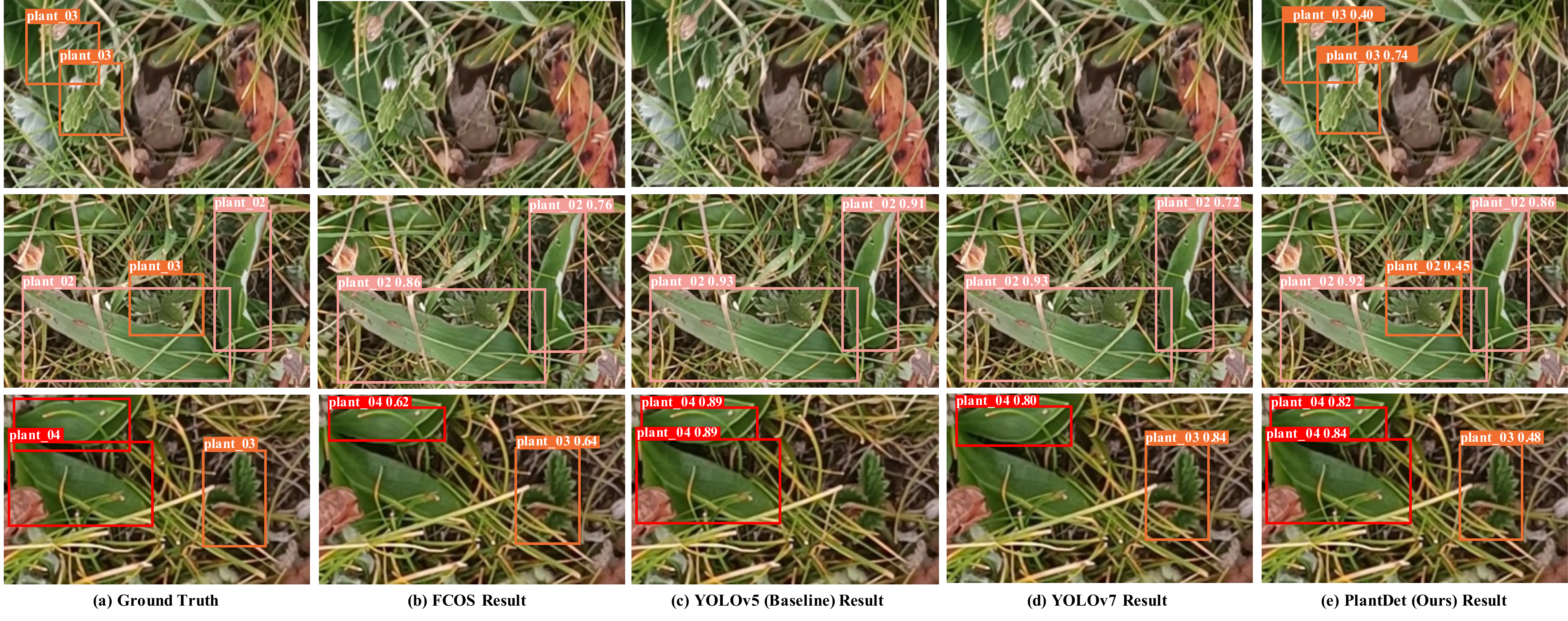}
\caption{Visualization results on our PTRS dataset.}
\label{fig:visualization}
\end{figure*}

\paragraph{\textbf{Qualitative Comparison.}} In order to further verify the superiority of our proposed PlantDet for plant detection, we conducted qualitative experiments to compare the detection performance of PlantDet and other models (FCOS, YOLOv5 and YOLOv7). The specific visualization results are shown in Fig. ~\ref{fig:visualization}. 

The visualization results 
show that compering with other models, our PlantDet can effectively prevent the occurrence of missed inspections and reduce the false detection rate while ensuring detection accuracy. In summary, our PlantDet has better performance for plant detection in the Three-River-Source region, and can meets the needs of botanical Studies and intelligent plant management.

\section{Conclusion}

To address the problem of varying leaf resolution, severe occlusion, and high feature similarity in plant species, we proposed a novel object detection benchmark called PlantDet. Experimental results show that our PlantDet achieves SOTA detection performance and effectively prevents false detection and missed detection. In addition, we construct a large-scale dataset for plant detection in the Three-River-Source region, which provides data foundation and technical support for the informatization of grassland resources and the construction of a smart ecological animal husbandry new model of "reducing pressure and increasing efficiency" for ecological protection of the Three-Rivers-Source region.
\section{Acknowledgments}
This study is supported by the Science and Technology Plan of Qinghai Province (2020-QY-218), and China Agriculture Research System of MOF and MARA (CARS-37).

%
%
%
\bibliographystyle{splncs04}
\bibliography{egbib}
%





\end{document}